\documentclass[runningheads]{llncs}
\usepackage{graphicx}
\usepackage{multirow}
\usepackage{color}
\begin{document}
\title{DeformableFormer: Classification of Endoscopic Ultrasound Guided Fine Needle Biopsy in Pancreatic Diseases
}



\author{Taiji Kurami\inst{1}\orcidID{0009-0002-8046-2362} \and
Takuya Ishikawa\inst{2}\orcidID{0000-0001-5814-3555} \and
Kazuhiro Hotta\inst{3}\orcidID{0000-0002-5675-8713}}
\authorrunning{F. Author et al.}
%
\institute{Meijo University, Aichi, Japan\\
\email{190442060@ccalumni.meijo-u.ac.jp}\and
Nagoya University, Aichi, Japan\\
\email{ishitaku@med.nagoya-u.ac.jp} \and
Meijo University, Aichi, Japan\\
\email{kazuhotta@meijo-u.ac.jp}}

\maketitle              
\begin{abstract}
Endoscopic Ultrasound-Fine Needle Aspiration (EUS-FNA) is used to examine pancreatic cancer. EUS-FNA is an examination using EUS to insert a thin needle into the tumor and collect pancreatic tissue fragments. Then collected pancreatic tissue fragments are then stained to classify whether they are pancreatic cancer. However, staining and visual inspection are time consuming. In addition, if the pancreatic tissue fragment cannot be examined after staining, the collection must be done again on the other day. 
Therefore, our purpose is to classify from an unstained image whether it is available for examination or not, and to exceed the accuracy of visual classification by specialist physicians. Image classification before staining can reduce the time required for staining and the burden of patients.
However, the images of pancreatic tissue fragments used in this study cannot be successfully classified by processing the entire image because the pancreatic tissue fragments are only a part of the image.
Therefore, we propose a DeformableFormer that uses Deformable Convolution in MetaFormer framework. The architecture consists of a generalized model of the Vision Transformer, and we use Deformable Convolution in the TokenMixer part. In contrast to existing approaches, our proposed DeformableFormer is possible to perform feature extraction more locally and dynamically by Deformable Convolution. Therefore, it is possible to perform suitable feature extraction for classifying target. 
To evaluate our method, we classify two categories of pancreatic tissue fragments; available and unavailable for examination. We demonstrated that our method outperformed the accuracy by specialist physicians and conventional methods.

\end{abstract}
\section{Introduction}

Currently, EUS-FNA is used to examine pancreatic cancer. It is an examination using EUS to insert a thin needle into the tumor and collect pancreatic tissue fragments. Then collected pancreatic tissue fragments are then stained to classify whether they are pancreatic cancer. However, staining and visual inspection are time consuming. After staining, if it is determined that the pancreatic tissue fragment cannot be examined because it has not been acquired sufficiently, the acquisition must be done on the other day. This is a big problem that increases the burden on the patient.
Therefore, it is desirable to be able to classify them in the state before staining, but it is difficult even for medical specialists to classify whether pancreatic tissue fragments have been acquired sufficiently before staining or not. 
Therefore, the objective of this study is to exceed the accuracy of classification by medical specialists using machine learning on before staining images.

In recent years, Transformer~\cite{Transformer1,Transformer2,Transformer3,Transformer4} has shown high accuracy in computer vision. Since Vision Transformer (ViT)~\cite{ViT} which uses a simple Transformer for image classification was introduced, various models~\cite{ViT_application1,ViT_application2,ViT_application3} have been developed to achieve higher accuracy in various tasks such as image classification~\cite{Classification1,Classification2}, object detection~\cite{detection5,detection1,detection2,detection3,detection4} and image generation~\cite{generation1,generation3,generation2}.

 \begin{figure}
     \centering
     \includegraphics[scale=0.5]{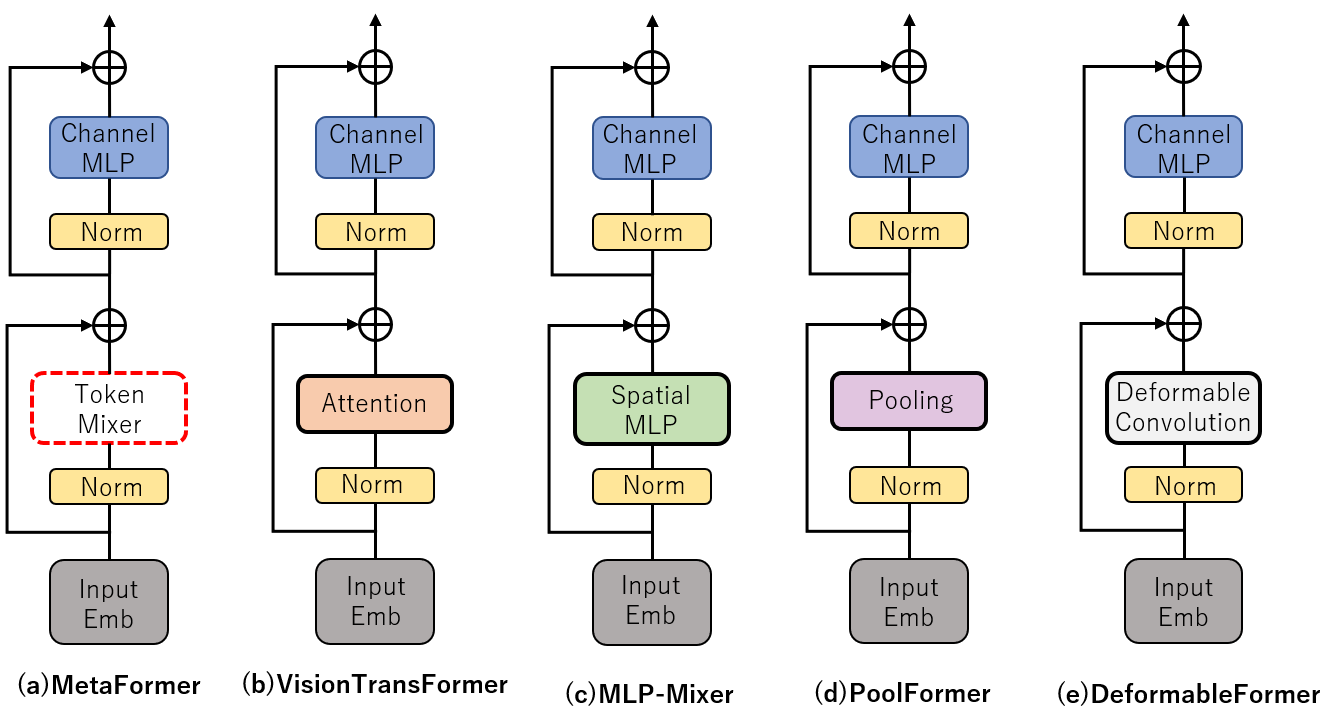}
     \caption{(a) MetaFormer~\cite{MetaFormer}: A model that generalizes Vision Transformer~\cite{ViT}. (b) Vision Transformer~\cite{ViT}: Token Mixing part is replaced by Attention. (c) MLP-Mixer~\cite{MLP-Mixer}: Token Mixing part is replaced by MLP. (d) PoolFormer~\cite{MetaFormer}: Token Mixing is replaced by Pooling. (e) The proposed DeformableFormer: The Token Mixing part is replaced by Deformable Convolution~\cite{DeformableConv}.}
     \label{fig:my_labe1}
 \end{figure}
 
MetaFormer~\cite{MetaFormer}, a model that generalizes Vision Transformer~\cite{ViT}, has been proposed. The reason why Transformer-based methods achieved high accuracy has been attributed to mix the information among tokens called TokenMixer as shown in Fig.~\ref{fig:my_labe1}. However, the PoolFormer~\cite{MetaFormer} which changed the TokenMixer part to Pooling and MLP-Mixer~\cite{MLP-Mixer} which changed the TokenMixer
part to MLP showed the similar performance. This suggests that the Transformer structure itself, not the TokenMixer part, was responsible for the performance.
 
We use MetaFormer~\cite{MetaFormer} as a baseline model. In addition, the images used in this study have many background areas and blood portions that are other than the pancreatic tissue fragment which is the classification target. Therefore, we propose DeformableFormer which incorporates Deformable Convolution~\cite{DeformableConv} into MetaFormer~\cite{MetaFormer}
so that features can be extracted from small pancreatic tissue fragments effectively. This enables the successful learning of the pancreatic tissue fragment to be classified. We realize the classification of before staining images which has been difficult even for medical specialists.

To validate the proposed method, we perform an image classification of two classes of pancreatic tissue fragments; available and unavailable for examination. When we compared our method with the accuracy of a medical specialist, the proposed method was improved to $5.20\%$ for the accuracy rate, $2.25\%$ for the precision rate, $4.13\%$ for the recall rate, and $10.72\%$ for the specificity rate.

The structure of this paper is as follows. Section 2 describes the proposed method. We show experimental results in Section 3. Finally, conclusion and future works are described in section 4.

\section{Proposed Method}

Conventional famous models such as Resnet~\cite{Resnet34} and PoolFormer~\cite{MetaFormer} perform Convolution and Pooling on the entire image with a defined kernel size. In addition, Vision Transformer~\cite{ViT} also processes the entire image using Self-Attention.
However, the images of pancreatic tissue fragments used in this study cannot be successfully classified by conventional static processing because the pancreatic tissue fragments are only a part of the image as shown in Figure 3. Therefore, we propose a DeformableFormer that incorporates Deformable Convolution into MetaFormer, which can perform Convolution according to the classification target. Detailed descriptions are provided in Sections 2.1 and 2.2.

\subsection{DeformableFormer}

\begin{figure}
     \centering
     \includegraphics[scale=0.4]{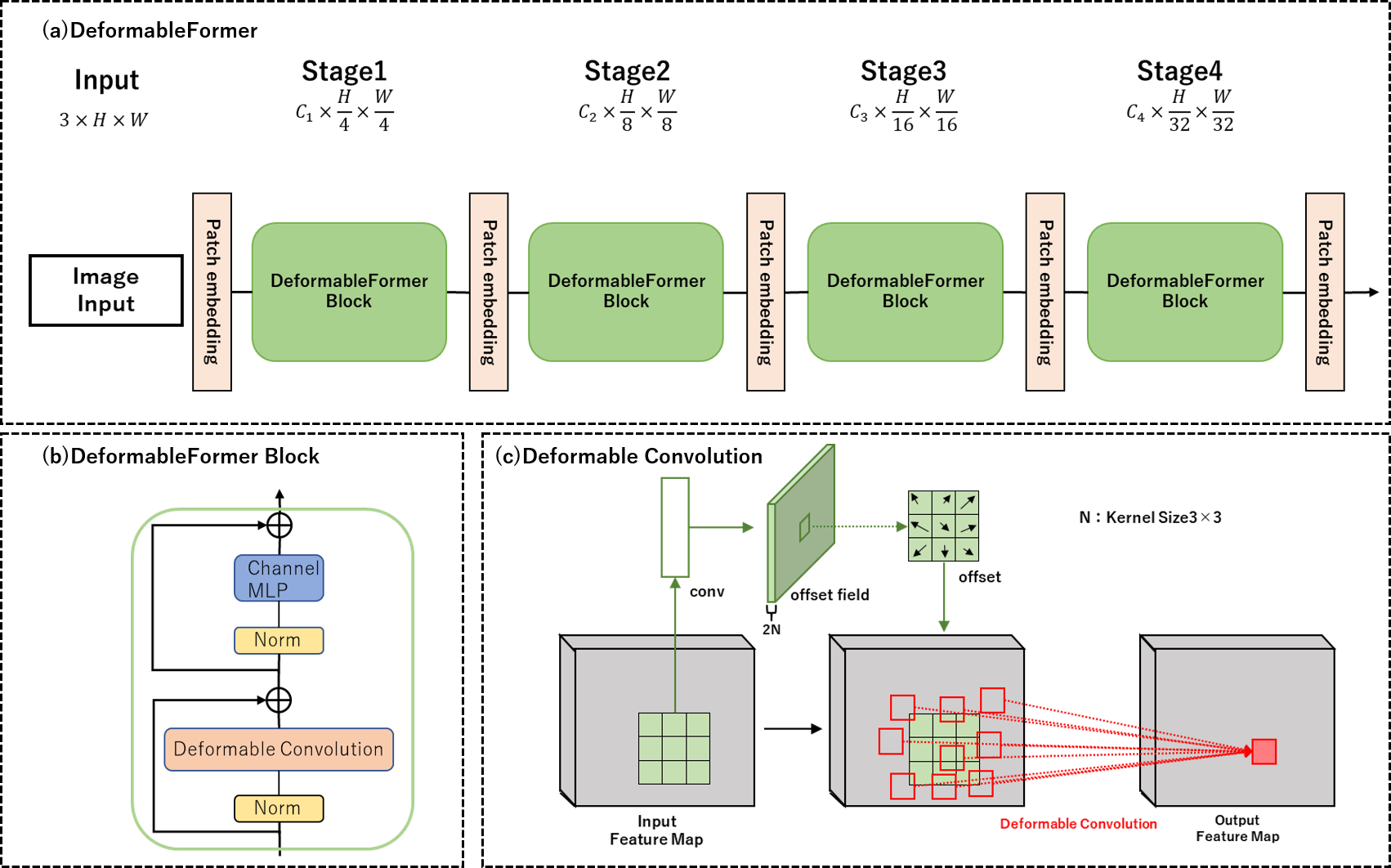}
     \caption{(a) Architecture of the DeformableFormer. (b) Architecture of DeformableFormer Block. (c) Architecture of Deformable Convolution~\cite{DeformableConv} in the Token Mixing of DeformableFormer Block.}
     \label{fig:my_labe2}
 \end{figure}

The architecture of the proposed DeformableFormer is shown in Fig.~\ref{fig:my_labe2}. In addition, Table~\ref{tab1} summarizes the image size, patch size, number of channels, kernel size, MLP Ratio, and number of blocks at each stage in DeformableFormer. From Fig.~\ref{fig:my_labe2} and Table~\ref{tab1}, DeformableFormer has a hierarchical structure in which the image size decreases from input to output, similar to CNN~\cite{CNN}. 
DeformableFormer first decomposes the image into patches and performs Deformable Convolution in the DeformableFormer Block at each stage. By using Deformable Convolution as Token Mixer, it is  expected that we can extract features according to the classification target.
We believe that this allows for more localized and dynamic feature extraction, and thus allows for successful learning and classification of even small objects such as pancreatic tissue fragments.

\begin{table}
\caption{Details of DeformableFormer architecture.}\label{tab1}
\centering
\begin{tabular}{|c|c|cc|c|}
\hline
Stage              & size                         & \multicolumn{2}{c|}{Layer}                                                                                                 & DeformableFormer \\ \hline
\multirow{5}{*}{1} & \multirow{5}{*}{$\frac{H}{4}$ $\times$ $\frac{W}{4}$}   & \multicolumn{1}{c|}{\multirow{2}{*}{\begin{tabular}[c]{@{}c@{}}Patch\\      Embedding\end{tabular}}}        & Patch Size   & 7×7, stride 4    \\ \cline{4-5} 
                   &                              & \multicolumn{1}{c|}{}                                                                                       & Channel  & 64               \\ \cline{3-5} 
                   &                              & \multicolumn{1}{c|}{\multirow{3}{*}{\begin{tabular}[c]{@{}c@{}}DeformableFormer\\      Block\end{tabular}}} & Kernel Size & 3×3, stride 1    \\ \cline{4-5} 
                   &                              & \multicolumn{1}{c|}{}                                                                                       & MLP Ratio    & 4                \\ \cline{4-5} 
                   &                              & \multicolumn{1}{c|}{}                                                                                       & Block        & 2                \\ \hline
\multirow{5}{*}{2} & \multirow{5}{*}{$\frac{H}{8}$ $\times$ $\frac{W}{8}$}   & \multicolumn{1}{c|}{\multirow{2}{*}{\begin{tabular}[c]{@{}c@{}}Patch\\      Embedding\end{tabular}}}        & Patch Size   & 3×3, stride 2    \\ \cline{4-5} 
                   &                              & \multicolumn{1}{c|}{}                                                                                       & Channel  & 128              \\ \cline{3-5} 
                   &                              & \multicolumn{1}{c|}{\multirow{3}{*}{\begin{tabular}[c]{@{}c@{}}DeformableFormer\\      Block\end{tabular}}} & Kernel Size & 3×3, stride 1    \\ \cline{4-5} 
                   &                              & \multicolumn{1}{c|}{}                                                                                       & MLP Ratio    & 4                \\ \cline{4-5} 
                   &                              & \multicolumn{1}{c|}{}                                                                                       & Block        & 2                \\ \hline
\multirow{5}{*}{3} & \multirow{5}{*}{$\frac{H}{16}$ $\times$ $\frac{W}{16}$} & \multicolumn{1}{c|}{\multirow{2}{*}{\begin{tabular}[c]{@{}c@{}}Patch\\      Embedding\end{tabular}}}        & Patch Size   & 3×3, stride 2    \\ \cline{4-5} 
                   &                              & \multicolumn{1}{c|}{}                                                                                       & Channel  & 320              \\ \cline{3-5} 
                   &                              & \multicolumn{1}{c|}{\multirow{3}{*}{\begin{tabular}[c]{@{}c@{}}DeformableFormer\\      Block\end{tabular}}} & Kernel Size & 3×3, stride 1    \\ \cline{4-5} 
                   &                              & \multicolumn{1}{c|}{}                                                                                       & MLP Ratio    & 4                \\ \cline{4-5} 
                   &                              & \multicolumn{1}{c|}{}                                                                                       & Block        & 6                \\ \hline
\multirow{5}{*}{4} & \multirow{5}{*}{$\frac{H}{32}$ $\times$ $\frac{W}{32}$} & \multicolumn{1}{c|}{\multirow{2}{*}{\begin{tabular}[c]{@{}c@{}}Patch\\      Embedding\end{tabular}}}        & Patch Size   & 3×3, stride 2    \\ \cline{4-5} 
                   &                              & \multicolumn{1}{c|}{}                                                                                       & Channel  & 512              \\ \cline{3-5} 
                   &                              & \multicolumn{1}{c|}{\multirow{3}{*}{\begin{tabular}[c]{@{}c@{}}DeformableFormer\\      Block\end{tabular}}} & Kernel Size & 3×3, stride 1    \\ \cline{4-5} 
                   &                              & \multicolumn{1}{c|}{}                                                                                       & MLP Ratio    & 4                \\ \cline{4-5} 
                   &                              & \multicolumn{1}{c|}{}                                                                                       & Block        & 2                \\ \hline
\end{tabular}
\end{table}

\subsection{Details of DeformableFormer Block}

The DeformableFormer Block is based on the MetaFormer~\cite{MetaFormer}
model which is a generalized version of the Vision Transformer~\cite{ViT} model, and performs Deformable Convolution~\cite{DeformableConv} in Token Mixer part to extract features according to the classification target. 
Deformable Convolution~\cite{DeformableConv} performed in Token Mixer consists of two steps: 1) The pixels for convolution are dynamically changed by using an offset $\Delta p$ in terms of input features $x$. This allows the input data to be cut out irregularly. Therefore, the convolution position is $x(p_0 + p_n + \Delta p_n)$ because it is obtained from the input image. 2) We take the inner product of the cropped data and the weights $\omega$. This enables convolution according to the classification target. 

The formula for Deformable Convolution is shown in Equation (1).
\begin{equation}
y(p_0) = \sum_{p_n\in R}\omega(p_0)\cdot x(p_0 + p_n + \Delta p_n)
\end{equation}
As shown in Fig.~\ref{fig:my_labe2}, the size of the offset field is the same as that of the input feature map, and the number of channels 2N corresponds to the kernel size N.

\section{Experiments}

\subsection{Implementation Details and Evaluation Methods}

\begin{figure}
     \centering
     \includegraphics[scale=0.5]{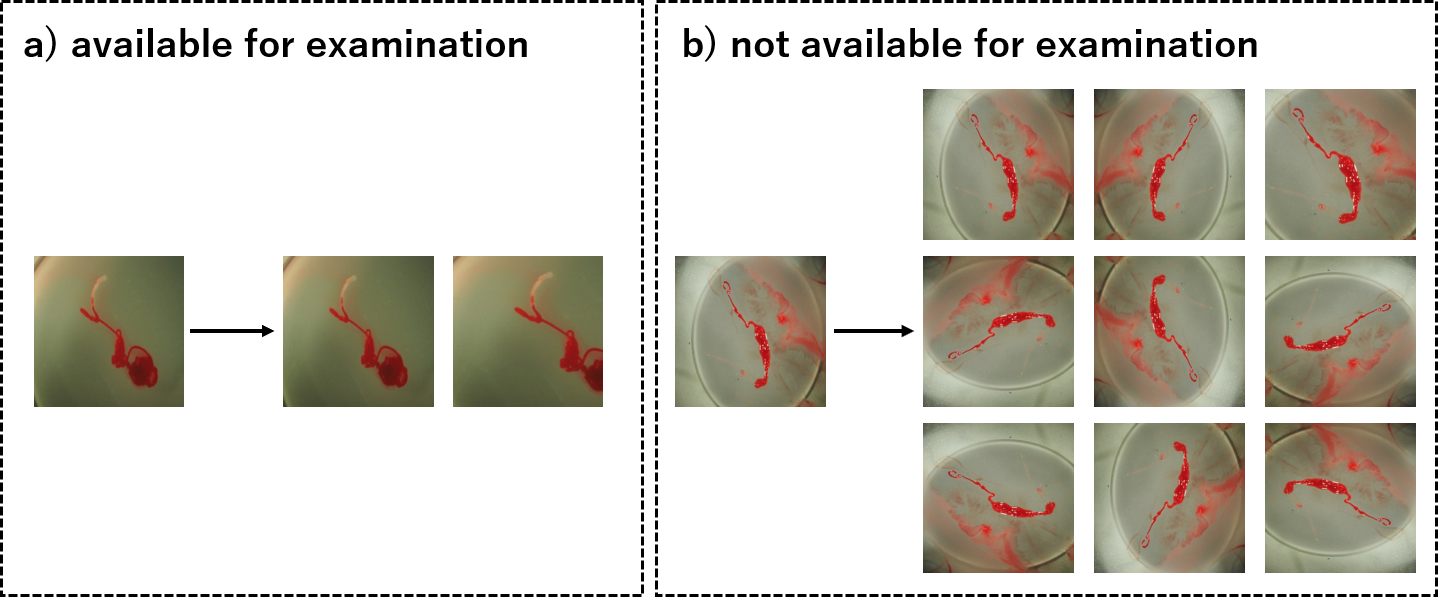}
     \caption{(a) The number of available images for examination is increased by a factor of two: the original image and the cropped image.  (b) The number of unavailable images for examination is increased by a factor of nine: the original image and the cropped image, plus the inverted image and the image rotated by 90 degrees.}
     \label{fig:my_labe3}
 \end{figure}

We use a pancreatic tissue fragment dataset provided by Nagoya University. The dataset consists of images before staining. We have two classes; $145$ images available for examination and $28$
images unavailable for examination. 
The number of images is increased by data augmentation because the number of training images (especially unavailable class) is small. Data augmentation for training images in two classes is shown in Fig.~\ref{fig:my_labe3}. By data augmentation shown in the Figure, the number of available images for examination is increased by a factor of two: the original image and the cropped image. The number of images that are not available for examination is increased by a factor of nine: the original image and the cropped image, plus the inverted image and the image rotated by 90 degrees.

We used cross validation because of the small number of images. 
All images are divided into $28$ sets, and $27$ sets are used for training and remaining $1$ set is used as test. Thus, $28$-fold cross-validation is used, in which one test data is shifted and we trained a model $28$ times. 
In addition, since the pixel size of the original image is $4608 \times 3456$ pixels, the original image is too big to compute with GPU. Thus, $1600 \times 1600$ resized image is used. This experiment was conducted with a batch size of 5, an epoch of 50, and a learning rate of 0.001.

Because this experiment is a binary classification of available and unavailable for examination, a confusion matrix is used. A confusion matrix is a table that summarizes the classification results and is used as a measure of the performance of binary classification.
We use four evaluation measures computed from a confusion matrix; accuracy rate, precision rate, recall rate, and specificity rate.

Accuracy rate is a measure of how well the overall prediction result matches the true value. The higher value is better. 
Therefore, the accuracy rate indicates the classification accuracy of two classes; available and unavailable for examination.
The formula is shown in Equation (2).
\begin{equation}
Accuracy = (TP+TN)/(TP+FP+FN+TN)
\end{equation}
where TP is true positive, TN is true negative, FP is false positive and FN is false negative. 
 
Precision rate indicates the percentage of true positive in samples predicted as positive. Therefore, the precision rate indicates the accuracy of available images for examination. If there are many false positives, precision rate decreases. 
The formula is shown in Equation (3).
\begin{equation}
Precision = TP/(TP+FP)
\end{equation}
 
Recall rate is the percentage of true positive in positive samples. Therefore, the recall rate also indicates the accuracy of available images for examination. If there are false negatives, recall rate decreases.
The formula is shown in Equation (4).
 \begin{equation}
Recall = TP/(TP+FN)
\end{equation}
 
Specificity rate is the percentage of true negative in unavailable images for examination. In other words, the specificity rate indicates whether unavailable class for examination is correctly classified. 
If unavailable images for examination are mis-classified as positive, 
then pancreatic tissue fragment cannot be examined after staining, and acquisition of  tissue fragment must be done again on the other day. 
Since this is the worse case, 
our primary goal is to improve the value of specificity rate.
The formula is shown in Equation (5). 
 \begin{equation}
 Specificity = TN/(FP+TN)
\end{equation}
 
In this experiment, it is difficult to classify whether a pancreatic tissue fragment is available or unavailable for examination when the amount of tissue fragment is small. In particular, it was very difficult even for medical specialists to classify the unavailable samples for examination before staining images.

\subsection{Comparison results}

\begin{table}
\caption{Confusion matrices by medical specialties, existing methods, and the proposed method. 
Available means images that are available for examination, and Unavailable means images that are not available for examination. The number means the number of images classified.}\label{tab3}
\centering
\begin{tabular}{cccclclcc}
\cline{1-4} \cline{6-9}
\multicolumn{2}{|c|}{\multirow{2}{*}{Medical   Specialist}}                      & \multicolumn{2}{c|}{Prediction}                                 & \multicolumn{1}{l|}{} & \multicolumn{2}{c|}{\multirow{2}{*}{Resnet34~\cite{Resnet34}}}                                  & \multicolumn{2}{c|}{Prediction}                                 \\ \cline{3-4} \cline{8-9} 
\multicolumn{2}{|c|}{}                                                           & \multicolumn{1}{c|}{Available} & \multicolumn{1}{c|}{Unavailable} & \multicolumn{1}{l|}{} & \multicolumn{2}{c|}{}                                                           & \multicolumn{1}{c|}{Available} & \multicolumn{1}{c|}{Unavailable} \\ \cline{1-4} \cline{6-9} 
\multicolumn{1}{|c|}{\multirow{2}{*}{True Label}} & \multicolumn{1}{c|}{Available}   & \multicolumn{1}{c|}{129}      & \multicolumn{1}{c|}{16}         & \multicolumn{1}{l|}{} & \multicolumn{1}{c|}{\multirow{2}{*}{True Label}} & \multicolumn{1}{c|}{Available}   & \multicolumn{1}{c|}{120}      & \multicolumn{1}{c|}{25}         \\ \cline{2-4} \cline{7-9} 
\multicolumn{1}{|c|}{}                         & \multicolumn{1}{c|}{Unavailable} & \multicolumn{1}{c|}{13}       & \multicolumn{1}{c|}{15}         & \multicolumn{1}{l|}{} & \multicolumn{1}{c|}{}                         & \multicolumn{1}{c|}{Unavailable} & \multicolumn{1}{c|}{19}       & \multicolumn{1}{c|}{9}          \\ \cline{1-4} \cline{6-9} 
\multicolumn{1}{l}{}                           & \multicolumn{1}{l}{}            & \multicolumn{2}{c}{}                                            &                       & \multicolumn{1}{l}{}                          &                                 & \multicolumn{2}{c}{}                                            \\ \cline{1-4} \cline{6-9} 
\multicolumn{2}{|c|}{\multirow{2}{*}{Contrastive   Learning~\cite{ishikawa2022development}}}                    & \multicolumn{2}{c|}{Prediction}                                 & \multicolumn{1}{l|}{} & \multicolumn{2}{c|}{\multirow{2}{*}{PoolFormer~\cite{MetaFormer}}}                                & \multicolumn{2}{c|}{Prediction}                                 \\ \cline{3-4} \cline{8-9} 
\multicolumn{2}{|c|}{}                                                           & \multicolumn{1}{c|}{Available} & \multicolumn{1}{c|}{Unavailable} & \multicolumn{1}{l|}{} & \multicolumn{2}{c|}{}                                                           & \multicolumn{1}{c|}{Available} & \multicolumn{1}{c|}{Unavailable} \\ \cline{1-4} \cline{6-9} 
\multicolumn{1}{|c|}{\multirow{2}{*}{True Label}} & \multicolumn{1}{c|}{Available}   & \multicolumn{1}{c|}{131}      & \multicolumn{1}{c|}{14}         & \multicolumn{1}{l|}{} & \multicolumn{1}{c|}{\multirow{2}{*}{True Label}} & \multicolumn{1}{c|}{Available}   & \multicolumn{1}{c|}{136}      & \multicolumn{1}{c|}{9}          \\ \cline{2-4} \cline{7-9} 
\multicolumn{1}{|c|}{}                         & \multicolumn{1}{c|}{Unavailable} & \multicolumn{1}{c|}{13}       & \multicolumn{1}{c|}{15}         & \multicolumn{1}{l|}{} & \multicolumn{1}{c|}{}                         & \multicolumn{1}{c|}{Unavailable} & \multicolumn{1}{c|}{14}       & \multicolumn{1}{c|}{14}         \\ \cline{1-4} \cline{6-9} 
\multicolumn{1}{l}{}                           & \multicolumn{1}{l}{}            & \multicolumn{1}{l}{}          & \multicolumn{1}{l}{}            &                       & \multicolumn{1}{l}{}                          &                                 & \multicolumn{1}{l}{}          & \multicolumn{1}{l}{}            \\ \cline{1-4}
\multicolumn{2}{|c|}{\multirow{2}{*}{Proposed method}}                                       & \multicolumn{2}{c|}{Prediction}                                 &                       & \multicolumn{1}{l}{}                          &                                 & \multicolumn{1}{l}{}          & \multicolumn{1}{l}{}            \\ \cline{3-4}
\multicolumn{2}{|c|}{}                                                           & \multicolumn{1}{c|}{Available} & \multicolumn{1}{c|}{Unavailable} &                       & \multicolumn{1}{l}{}                          &                                 & \multicolumn{1}{l}{}          & \multicolumn{1}{l}{}            \\ \cline{1-4}
\multicolumn{1}{|c|}{\multirow{2}{*}{True Label}} & \multicolumn{1}{c|}{Available}   & \multicolumn{1}{c|}{135}      & \multicolumn{1}{c|}{10}         &                       & \multicolumn{1}{l}{}                          &                                 & \multicolumn{1}{l}{}          & \multicolumn{1}{l}{}            \\ \cline{2-4}
\multicolumn{1}{|c|}{}                         & \multicolumn{1}{c|}{Unavailable} & \multicolumn{1}{c|}{10}       & \multicolumn{1}{c|}{18}         &                       & \multicolumn{1}{l}{}                          &                                 & \multicolumn{2}{c}{}                                            \\ \cline{1-4}
\end{tabular}
\end{table}

\begin{table}
\caption{Accuracy at four evaluation measures. Increase Rate is the degree of increase compared to the accuracy of medical specialists. Bold letters indicate the highest accuracy in all methods.}\label{tab4}
\centering
\begin{tabular}{|c|cc|cc|}
\hline
\multirow{2}{*}{}    & \multicolumn{2}{c|}{Accuracy   rate}                    & \multicolumn{2}{c|}{Precision   rate}                    \\ \cline{2-5} 
                     & \multicolumn{1}{c|}{Accuracy}         & Increase Rate   & \multicolumn{1}{c|}{Accuracy}         & Increase Rate    \\ \hline
Medical Specialist   & \multicolumn{1}{c|}{83.24\%}          & -               & \multicolumn{1}{c|}{90.85\%}          & -                \\ \hline
Resnet34~\cite{Resnet34}             & \multicolumn{1}{c|}{74.57\%}          & -8.67\%         & \multicolumn{1}{c|}{86.33\%}          & -4.52\%          \\ \hline
Contrastive Learning~\cite{ishikawa2022development} & \multicolumn{1}{c|}{84.39\%}          & 1.15\%          & \multicolumn{1}{c|}{90.97\%}          & 0.12\%           \\ \hline
PoolFormer~\cite{MetaFormer}           & \multicolumn{1}{c|}{86.71\%}          & 3.47\%          & \multicolumn{1}{c|}{90.67\%}          & -0.18\%          \\ \hline
Proposed method                  & \multicolumn{1}{c|}{\textbf{88.44\%}} & \textbf{5.20\%} & \multicolumn{1}{c|}{\textbf{93.10\%}} & \textbf{2.25\%}  \\ \hline \hline
\multirow{2}{*}{}    & \multicolumn{2}{c|}{Recall rate}                        & \multicolumn{2}{c|}{Specificity rate}                    \\ \cline{2-5} 
                     & \multicolumn{1}{c|}{Accuracy}         & Increase Rate   & \multicolumn{1}{c|}{Accuracy}         & Increase Rate    \\ \hline
Medical Specialist   & \multicolumn{1}{c|}{88.97\%}          & -               & \multicolumn{1}{c|}{53.57\%}          & -                \\ \hline
Resnet34~\cite{Resnet34}             & \multicolumn{1}{c|}{82.76\%}          & -6.21\%         & \multicolumn{1}{c|}{32.14\%}          & -21.43\%         \\ \hline
Contrastive Learning~\cite{ishikawa2022development} & \multicolumn{1}{c|}{90.34\%}          & 1.37\%          & \multicolumn{1}{c|}{53.57\%}          & 0.00\%           \\ \hline
PoolFormer~\cite{MetaFormer}           & \multicolumn{1}{c|}{\textbf{93.79\%}} & \textbf{4.82\%} & \multicolumn{1}{c|}{50.00\%}          & -3.57\%          \\ \hline
Proposed method                 & \multicolumn{1}{c|}{93.10\%}          & 4.13\%          & \multicolumn{1}{c|}{\textbf{64.29\%}} & \textbf{10.72\%} \\ \hline
\end{tabular}
\end{table}

This section shows comparison results with medical specialists and conventional methods. We used two conventional methods; Resnet34~\cite{Resnet34} used in \cite{ishikawa2022development}, the method using contrastive learning~\cite{ishikawa2022development}
that brings the features between pre-stained images and post-stained images with the same class closer.
We also evaluate PoolFormer~\cite{MetaFormer} which is the baseline of our method. 

Table~\ref{tab3} shows the confusion matrices of each method. Available in Table indicates available images for examination, and Unavailable means unavailable images for examination. The number in the Table indicates the number of images classified. 

Table~\ref{tab4} shows the accuracy at four evaluation measures. 
Accuracy shows the accuracy of each of four evaluation measures, and Increase Rate shows the rate of increase in the accuracy of each method compared to the accuracy of the medical specialist. In addition, bold letters indicate the best accuracy.

Table~\ref{tab4} shows that the method 
using contrastive learning~\cite{ishikawa2022development} improved the accuracy in the three evaluation measures in comparison with medical specialist. 
However, the accuracy of the specificity rate were not improved.
This indicates that the method using constrastive learning is not able to classify unavailable images for examination well. 
In addition, PoolFormer improved the accuracy in two evaluation measures in comparison with medical specialist, but pecision rate and specificity rate decreased. Because the contrastive learning based method~\cite{ishikawa2022development} and PoolFormer~\cite{MetaFormer} perform feature extraction at a fixed size, such as Convolution and Pooling, it is considered that the accuracy is reduced when the amount of pancreatic tissue fragments to be classified is small, because the features cannot be extracted well.

In contrast, the proposed method improved the accuracy at all four evaluation measures in comparison with medical specialists. Therefore, the proposed method shows overall higher performance than other methods. In particular, the best specificity rate by our method demonstrated that it is highly effective for classifying images that are not available for examination, which is the primary goal of the study.

\section{Conclusion}

The proposed method improved the accuracy of both classes compared to medical specialists and conventional methods. 
In particular, when we pay attention to the specificity rate which is the most important measure, Resnet34 and PoolFormer decreased the accuracy by $-21.43\%$ and $-3.57\%$ in comparison with the medical specialist. The conventional method using contrastive learning did not improve the specificity rate.
However, the proposed DeformableFormer improved the specificity rate by $10.72\%$ compared to the accuracy of the specialists. An increase in the specificity rate means an improvement in the classification of images that are not available for examination. This is because the DeformableFormer performs feature extraction according to the classification target which successfully learns the pancreatic tissue fragments.
Thus, our method improved the classification accuracy even for images that are not available for examination, for which training data is scarce.

Although our method improved the specificity rate, it is not still high.
There is also the issue of a small number of training images, which prevents the accuracy from improving. Therefore, it is necessary to address the issue in the future.

%
%
%
\bibliographystyle{splncs04}
\bibliography{mybibliography}
%




\end{document}